\documentclass{article}

\usepackage{times}
\usepackage{graphicx} 
\usepackage{subfigure} 

\usepackage{natbib}

\usepackage{algorithm}
\usepackage{algorithmic}

\usepackage{hyperref}

\usepackage[accepted]{icml2017} 

\usepackage{lipsum}

\usepackage{csquotes}

\icmltitlerunning{On the Robustness of Convolutional Neural Networks to Internal Architecture and Weight Perturbations}

\begin{document} 

\twocolumn[
\icmltitle{On the Robustness of Convolutional Neural Networks\\to Internal Architecture and Weight Perturbations}




\icmlsetsymbol{equal}{*}

\begin{icmlauthorlist}
\icmlauthor{Nicholas Cheney}{equal,cornell}
\icmlauthor{Martin Schrimpf}{equal,se,harvard}
\icmlauthor{Gabriel Kreiman}{harvard}
\end{icmlauthorlist}

\icmlaffiliation{cornell}{Cornell University, Ithaca, New York, USA}
\icmlaffiliation{se}{Program in Software Engineering, Augsburg University and Technische Universit\"at M\"unchen and Ludwig-Maximilians-Universit\"at M\"unchen, Germany}
\icmlaffiliation{harvard}{Boston Children's Hospital, Harvard Medical School, Boston, Massachusetts, USA}

\icmlcorrespondingauthor{Gabriel Kreiman}{gabriel.kreiman@childrens.harvard.edu}

\icmlkeywords{convolutional, deep, neural networks, machine learning, topology, architecture, mutation, sensitivity to noise, ICML}

\vskip 0.3in
]


\printAffiliationsAndNotice{\icmlEqualContribution} 

\begin{abstract}
Deep convolutional neural networks are generally regarded as robust function approximators. So far, this intuition is based on perturbations to external stimuli such as the images to be classified.  Here we explore the robustness of convolutional neural networks to perturbations to the internal weights and architecture of the network itself.  We show that convolutional networks are surprisingly robust to a number of internal perturbations in the higher convolutional layers but the bottom convolutional layers are much more fragile.
For instance, Alexnet shows less than a 30\% decrease in classification performance when randomly removing over 70\% of weight connections in the top convolutional or dense layers but performance is almost at chance with the same perturbation in the first convolutional layer.
Finally, we suggest further investigations which could continue to inform the robustness of convolutional networks to internal perturbations.  
\end{abstract}
%
\section{Introduction}

Current deep learning techniques are able to learn rich feature representations from large datasets with a general-purpose learning procedure by adjusting their internal parameters~\cite{lecun2015deep}.
These weights in neural networks are generally fit to the data once during training and then kept fixed for testing.

Online learning algorithms, in which data become gradually available and are presented sequentially, may be preferable to these batch algorithms, which learn on a dedicated training set, as the size of datasets scale~\cite{lecun2012efficient,bottou2003large} or if the data generating source changes over time, relaxing the assumption of independent and identically distributed random variables.

Unsupervised (or semi-supervised) deep learning algorithms are also highly desirable~\cite{salakhutdinov2009deep,lee2009convolutional,bengio2012unsupervised}, as they allow neural networks to approach problems that do not have an abundance of labeled data and vastly scale their real world applicability.  

However, online learning (and especially unsupervised online learning)
methods applied in practice may result in unpredictable changes to a deep neural network's architecture (potentially both topology and weights) during its real time use.  Especially in high-risk scenarios -- such as driverless cars~\cite{huval2015empirical}, UAVs~\cite{ross2013learning}, or autonomous robotics~\cite{levine2016end} -- it is important to understand how internal changes to a deep network during its execution will impact its immediate performance.

While weight changes to synapses represent experience-dependent changes during execution,
the performance effects of topological changes are also an area of concern.  Embodied machines, especially those with plastic morphologies such as modular robotics~\cite{yim2007modular}, may benefit from neurogenesis~\cite{chiel1997brain}.  Conversely, topological (or weight) changes could stem from defects in hardware implementations of neural networks~\cite{mead2012analog,schneider2017deeper}.

In biological systems, connectivity and weights between neurons in the brain are continuously changing. Neurons die every day and, at least in some parts of the brain, new neurons are formed~\cite{cunningham1982naturally,eriksson1998neurogenesis}. 
Synapses are subject to learning mechanisms -- such as spike-timing dependent plasticity~\cite{song2000competitive,hebb2005organization} -- that directly and dynamically modify the magnitude of connection strengths. 
It is remarkable that the brain is able to encode stable information in the presence of such drastic architectural changes. How brains simultaneously achieve plasticity and stability represents an open question in our understanding of biological information processing.

In the work reported here, we set forth to investigate the degree of robustness of information encoding (measured through image classification performance) under random destructive perturbations to the internal architecture (or topology) and weights of various deep convolutional neural network architectures.

\section{Background}

Regularizers like weight decay~\cite{Moody1991} or dropout~\cite{Srivastava2014} can be used during training to simulate weight changes with slowly decreasing parameter values and randomly dropped units respectively.
During test time, analyses like adversarial perturbations of the images~\cite{Goodfellow2014, Nguyen2015} and randomized labels~\cite{Zhang2016} have shed light onto the robustness of neural networks to changes in the data.

Learning curves of neural network performance during training are widely explored (including convolutional neural networks; e.g.~\cite{hinton2012improving}), as the performance increases asymptotically due to the intentional and directed internal parameter changes from learning.  

The robustness to perturbations to the internal architecture of neural networks have been studied in fully-connected neural networks~\cite{widrow199030}, including: perceptron networks~\cite{zurada1997perturbation}, Hopfield networks~\cite{liao1998robust}, recurrent neural networks~\cite{jabri1992weight}.

The current work extends this previous knowledge by focusing on quantifying the robustness of pre-trained deep convolutional networks to dynamic changes in the architecture and weights. 


%

\section{Methodology}

We mainly examine topology and weight perturbations to the deep convolutional neural network \textquote{Alexnet}~\cite{krizhevsky2012imagenet}, pretrained on the 2012 ImageNet Large Scale Visual Recognition Challenge (ILSVRC2012)~\cite{deng2009imagenet,russakovsky2015imagenet} but later also generalize results to VGG-16~\cite{simonyan2014very}.
This pretrained network is publicly available from the Caffe Zoo~\cite{jia2014caffe}.  
All models are implemented using Keras~\cite{Chollet2015} with a Theano backend~\cite{team2016theano}.

To perform each perturbation, a single layer of the network is chosen and the weights and biases (both from now on jointly referred to as \enquote{weights}) leading into that layer are separately modified according to one of the following procedures:
\begin{itemize}
\item {\it Synapse knockouts}. A proportion of weights leading into the given layer are set to zero.  These weights are randomly selected, as to evenly distribute the impact of the perturbation across all nodes in the layer.
\item {\it Node knockouts}. All of the weights leading into a proportion of nodes in the given layer are set to zero.  This has the effect of functionally removing those nodes from the network, and represents a more concentrated perturbation than the synapse knockout treatment.
\item {\it Weight perturbations}. Every weight leading into a given layer is increased (or decreased) by a random value drawn from a Gaussian distribution of a given spread (and centered around zero).  In this case, it is unlikely for any single synapse to be removed from the network, but instead all synapses will weight their incoming information slightly differently.  
\end{itemize}

Preliminary experiments applied perturbations to all layers of the network simultaneously. However, as evidenced in the results below, perturbations were found to have significantly different effects on classification performance when applied to different layers in the convolutional network.  Thus all experiments included below apply targeted perturbations to only one layer of the network at a time, and report the effect of that perturbation separately for each layer.  

We define performance as the proportion of images in the ILSVRC2012 validation set which the classifier correctly labels with one of its 5 most-highly-expressed output nodes (\textquote{top-5 Performance}). 



For all figures throughout this paper, values shown represent the mean classification performance of 3 to 5 runs with independent random perturbations, while error bars represent the standard deviation of the performance of these runs.  All $p$-values are calculated using the Wilcoxon rank-sum test~\cite{wilcoxon1945individual}. 


\section{Robustness to Topology Changes}


\subsection{Synapse Knockouts}

Perhaps the simplest topological change to a convolution neural network is the removal of a synapse.  This method randomly sets the weight of a proportion of synapses going into a given layer to zero, removing information uniformly and randomly from all nodes throughout that layer.  

\begin{figure}[t]
\vskip 0.2in
\begin{center}
\centerline{
\includegraphics[width=\columnwidth]{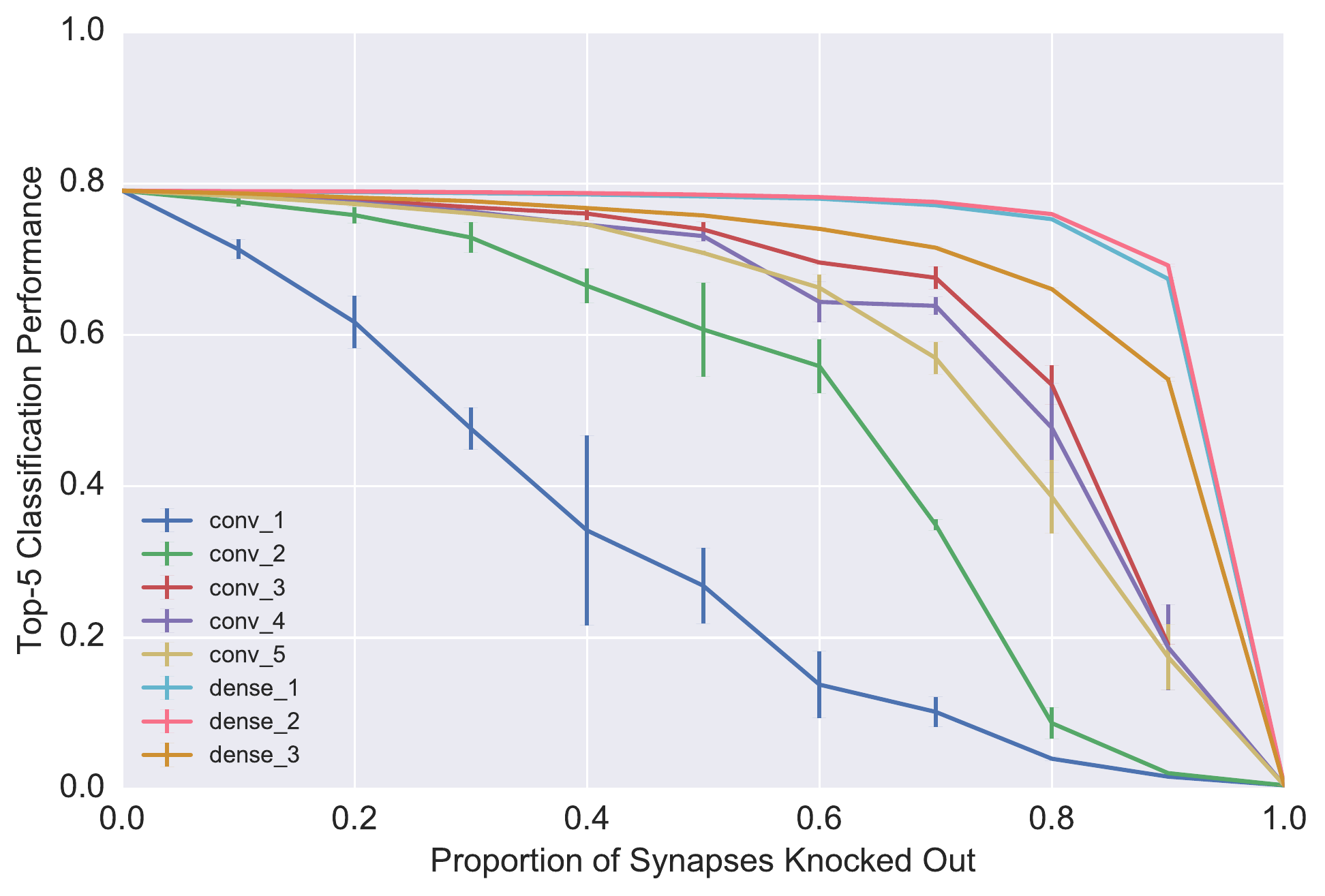}
}
\caption{Alexnet top-5 classification performance on ILSVRC after perturbing the weights of a given layer with \emph{synapse knockout}. Colors connote different convolutional layers, listed in order of appearance in the network. The $x$-axis denotes the proportion of synapses which are randomly removed for that layer.  Note the linear drop-off in the performance of the networks as the proportion of removed synapses in conv\_1 increases, but the non-linear relationship between synapse removal and performance in convolutional layers 2-5.}
\label{fig:perturbation:synapse-knockout}
\end{center}
\vskip -0.2in
\end{figure}

Fig.~\ref{fig:perturbation:synapse-knockout} shows the fall-off in classification performance due to knocking out an increasing proportion of synapses in different layers of the Alexnet convolutional neural network.  

Note that, for all layers, knocking out 0\% of synapses simply corresponds to the original unperturbed Alexnet, and thus results in its classification performance of 0.791.  At the other extreme, knocking out 100\% of any layer means that no information about a given input image is able to reach the output layer of the network, and thus the classification performance is no better than chance (which is 5 guesses / 1000 classes = 0.005 top-5 performance).

Classification performance for networks with synapse knockouts to the top 3 convolutional layers (conv\_3, conv\_4, and conv\_5) show remarkable robustness, loosing less than 10.5\% of their classification performance ability (from 0.791 to $\geq 0.708$) with up to 50\% of their synapses removed.  

Networks with 50\% of synapses removed from layer conv\_2 perform significantly worse than the top 3 convolutional layers ($p<0.05$), losing an average of 23.2\% of their classification performance (from 0.791 to 0.607).  Networks with 50\% of synapses removed from the first convolutional layer (conv\_1) perform significantly worse than the top 4 convolutional layers ($p<0.05$), losing an average of 66.0\% of their classification performance ability (from 0.791 to 0.269). 

The differences in sensitivity to synapse knockout perturbations in Fig.~\ref{fig:perturbation:synapse-knockout} is striking, showing much more fragility in lower layers.  For example, the removal of 30\% of the synapses in conv\_1 results in a network which performs significantly worse ($p<0.05$) than the network resulting from removal of 70\% of the synapses from any of the top 3 convolutional layers.  

The dense layers of Fig.~\ref{fig:perturbation:synapse-knockout} also demonstrate extreme robustness to synapse knockouts -- with dense\_1 and dense\_2 showing less than a 4.759\% drop in performance for knockouts of 80\% of synapses or less.  The robustness of these two layers to synapse knockout is unsurprising, as these layers are explicitly trained to minimize reliance on any single feature using dropout~\cite{srivastava2014dropout}.  


\subsection{Node Knockouts}

To further investigate topology changes to convolutional neural networks, we randomly removed a given number of nodes in each layer of the network. In the convolutional layers of a neural network, this corresponds to the removal of a convolutional filter.
Compared to the synapse knockout experiments in the previous section, node knockouts represent a more clustered and focused removal of information processing ability from the network.  

\begin{figure}[t]
\vskip 0.2in
\begin{center}
\centerline{
\includegraphics[width=\columnwidth]{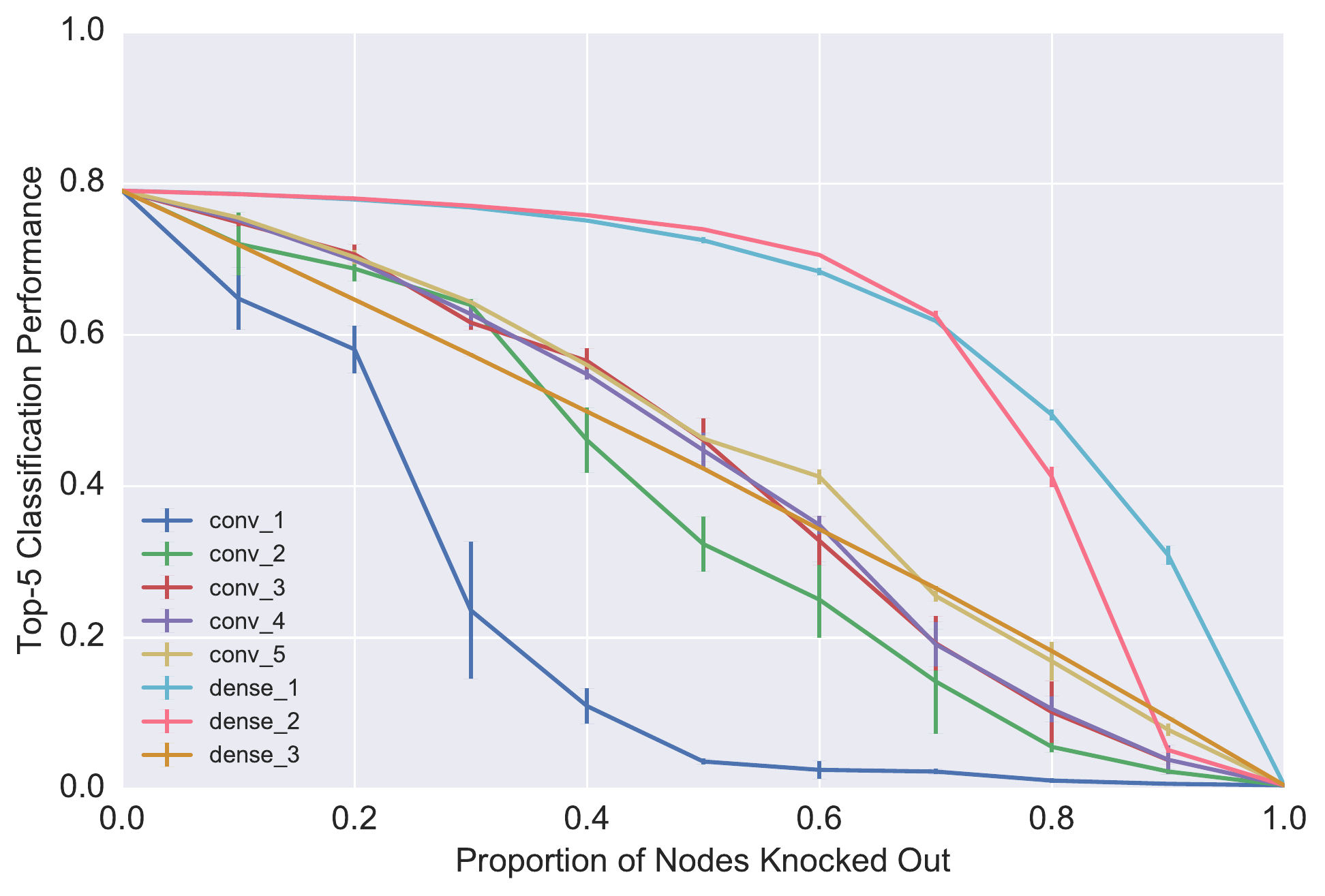}
}
\caption{Alexnet top-5 classification performance on ILSVRC after perturbing the weights of a given layer with \emph{node knockout}. The $x$-axis denotes the proportion of nodes which are randomly removed for that layer.  Note the linear relationship between networks performance and the proportion of nodes removed in convolutional layers 2-5, but the non-linear relationship between node removal and performance in conv\_1.}
\label{fig:perturbation:knockout}
\end{center}
\vskip -0.2in
\end{figure}

Fig.~\ref{fig:perturbation:knockout} shows the effect of node knockout on a given layer of Alexnet on the network's classification performance.  
As described in Fig.~\ref{fig:perturbation:synapse-knockout}, the effect of node knockout also revealed a large difference across layers. The network was significantly more labile to modifications in the earlier layers.

Since all nodes within a layer contain the same number of incoming synapses each, we are able to directly compare knockouts that are of the same \textquote{magnitude} (in terms of the total number of synapses removed) -- but differently distributed -- by comparing perturbations of the same proportion under the synapse knockout and node knockout treatments. 

Classification performance after knocking out a given proportion of nodes is significantly more damaging to the network than removing that same proportion of randomly distributed synapses.
This is true for networks perturbed with all combinations of intermediate knockout proportions (0.1 - 0.9) and layers (all $p<0.05$), with the exception of knocking out 90\% of nodes in the second convolutional layer -- which was not significantly different than knocking out a random 90\% of synapses ($p=0.827$).

One possible explanation for this may be that node knockout effectively reduces the size of a hidden layer, and that the function approximation abilities of a neural network can be constrained by its hidden layer size~\cite{hornik1989multilayer}.  

It may also be the case that information encoded in certain convolutional filters is unique to that filter and not always replicated in others (in the extreme case of this, the filters would represent orthogonal basis vectors over the space of features).  If this is the case, then removing that node would remove the information about its encoded \textquote{feature} from the neural network.  By \textquote{feature} we simply mean some unique and invariant property of the image.  This notion of \textquote{feature} may be fitting with our intuition of angled edges (for lower layers) or object parts (for upper layers) that are seen in filter activation visualizations~\cite{zeiler2014visualizing} -- but this need not be the case.     

In contrast to this, the removal of a synapse from a filter would corrode the ability of that filter to capture all of the information it previously did, but the remaining synapses may still provide a projection of the originally encoded information -- and perhaps a projection which is able to retain a greater proportion of its originally encoded information than the proportion of its original synapses which are left intact.  If this is the case, it would represent an especially robust property of neural networks to synapse knockout (at least compared to similar proportions of node knockout -- where the amount of previously encoded information that is removed may be proportional to the number of nodes that are knocked out).  

The idea that nodes represent \textquote{features}
-- especially in the higher level convolutional layers -- is further supported by the fact that classification performance falls off almost linearly as the proportion of nodes knocked out increases.  While this is the case in convolutional layers 2-5, as demonstrated by an r-squared value of 0.963 for a linear fit, this linear drop in performance is less evident in the first convolutional layer (where a linear fit gives an r-squared of 0.715). This observation should be contrasted with the fall-off in classification performance as the proportion of random synapses is removed (Fig.~\ref{fig:perturbation:synapse-knockout}), where higher convolutional layers (conv 2-5) demonstrate a less linear fit (r-squared of 0.553) and the first convolutional presents a linear decrease in classification performance as the proportion of randomly removed synapses increases (r-squared of 0.930).  

The knockout of nodes in dense layers (dense\_1 and dense\_2), display more robustness than the linear drop-off of the convolutional layers.  Again, the use of dropout allowed these layers to train in the presence of node knockouts, so the increased robustness here comes as no surprise.  The linear drop-off in the output layer (dense\_3) represents the notion that a class label becomes inaccessible if the output node corresponding to that class is knocked out.  Thus the perfectly linear drop-off in classification in response to output layer provides little information, but acts as a test case for our implementation of node knockouts.

\section{Robustness to Weight Changes from Random Perturbations}

Rather than removing synapses entirely and setting their weight to zero, both biological noise sources and computational applications  can lead to scenarios where perturbations to a network alter the weights by some amount. These weight modification situations are likely to be a concern in the case of unsupervised (or semi-supervised) learning rules where weights may change in an undesired (or simply random) direction, or in the case of damage or noise to embedded circuits. 
Such sources of noise or weight fluctuations may only cause a short term detriment to the performance of a networks -- but in critical use cases, even the characterization of short term effects may be of importance.  

In this treatment, \emph{all} of the synapses leading into a given layer are modified.  The size of this perturbation for each individual synapse is randomly drawn from a Gaussian distribution.  This Gaussian is centered at zero and has a standard deviation relative to the standard deviation of that layer's original weight distribution (e.g. a perturbation of magnitude 1 in the first convolutional layer corresponds to a change to each weight in that layer drawn from a Gaussian with a standard deviation equal to that of this first convolutional layer in the pretrained Alexnet).  The statistical properties of the original weight distributions are given in Table~\ref{alexnetTable} (located at the end of the text).  


This treatment differs from the case of synapse knockouts above, because all weights are affected and because the system may be led to put more or less emphasis on a given piece of incoming information, but that information is not entirely removed for the network -- as was the case in the treatments above.
  



\begin{figure}[ht]
\vskip 0.2in
\begin{center}
\centerline{\includegraphics[width=\columnwidth]{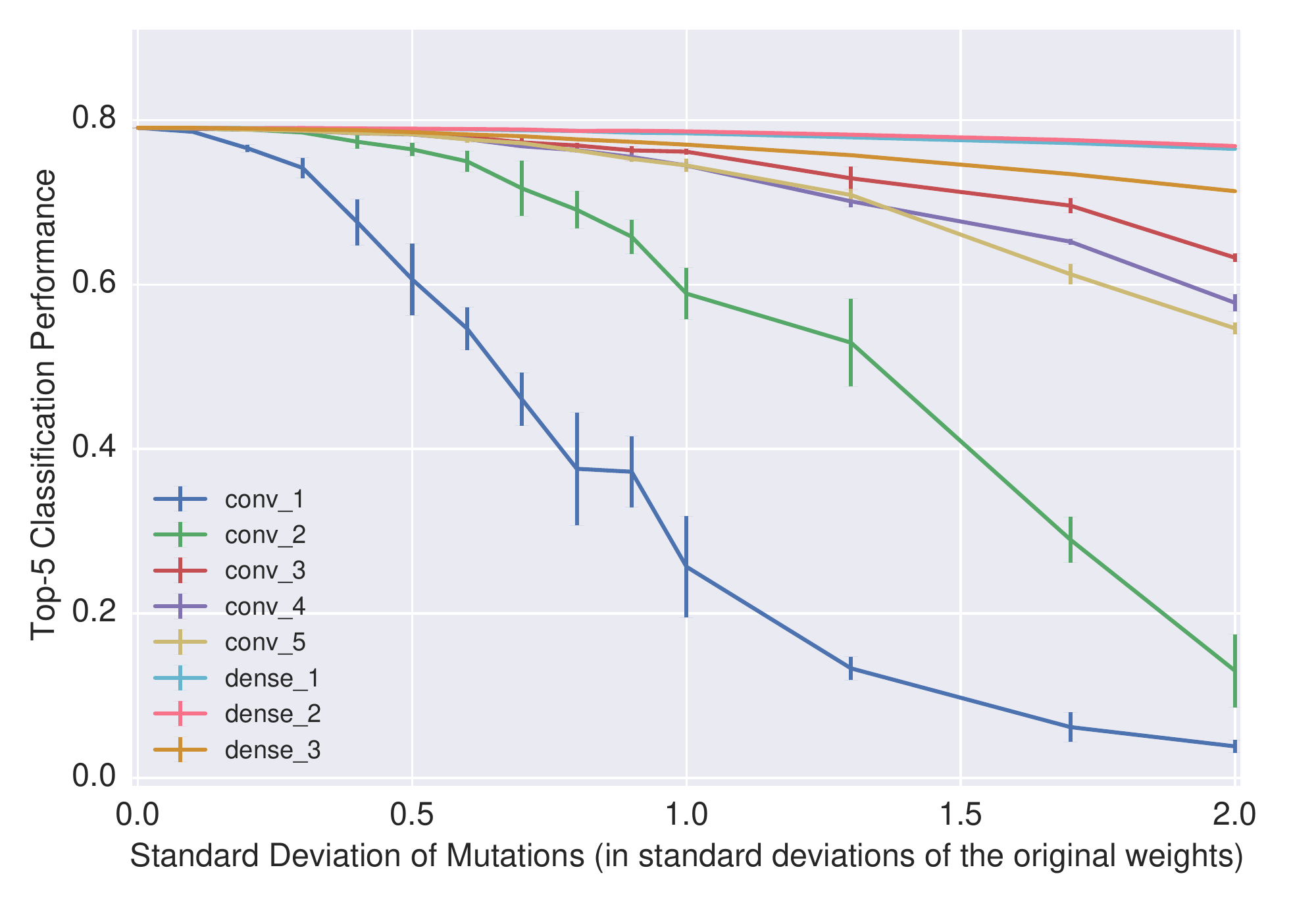}}
\caption{Alexnet top-5 performance on ILSVRC after perturbing the weights with Gaussian mutations.  The $x$-axis denotes the standard deviation of the distribution of mutation sizes, relative to the standard distribution of weights in that layer prior to perturbation.}
\label{fig:perturbation:mutate-performance}
\end{center}
\vskip -0.2in
\end{figure}





The drop in performance is generally more pronounced for lower convolutional layers.  In comparison between perturbations to different convolutional layers, the perturbations to the lower layer (i.e. closer to the input layer) resulted in worse performing networks  than those perturbed to a higher layer (all $p$-values $\leq$ 0.0495).  The exceptions were that perturbations to conv\_4 and conv\_5 were not worse that perturbations to convolutional layers above conv\_3 (and conv\_4).  


It is likely the case that the perturbation effects to the fully-connected dense layers are different that those to the convolutional layers, as their behaviors are fundamentally different.  In the convolutional layers, higher-level features are hierarchically composed from \textquote{sub-features} from layers below them -- and these features are applied spatially throughout the image with a sliding window.  Conversely, dense layers take these features and attempt to associate their presence or absence with the provided class labels.  As perturbations to fully connected neural network layers have been previously studied (Background), we focus largely on the convolutional layers.  

\section{Generalization to Other Metrics and Architectures}

While the top-5 classification performance of the Alexnet architecture on the ILSVRC2012 validation set has served as the focus for our initial studies, these results generalize well to various implementation decisions.  

\begin{figure}[ht]
\vskip 0.2in
\begin{center}
\centerline{\includegraphics[width=\columnwidth]{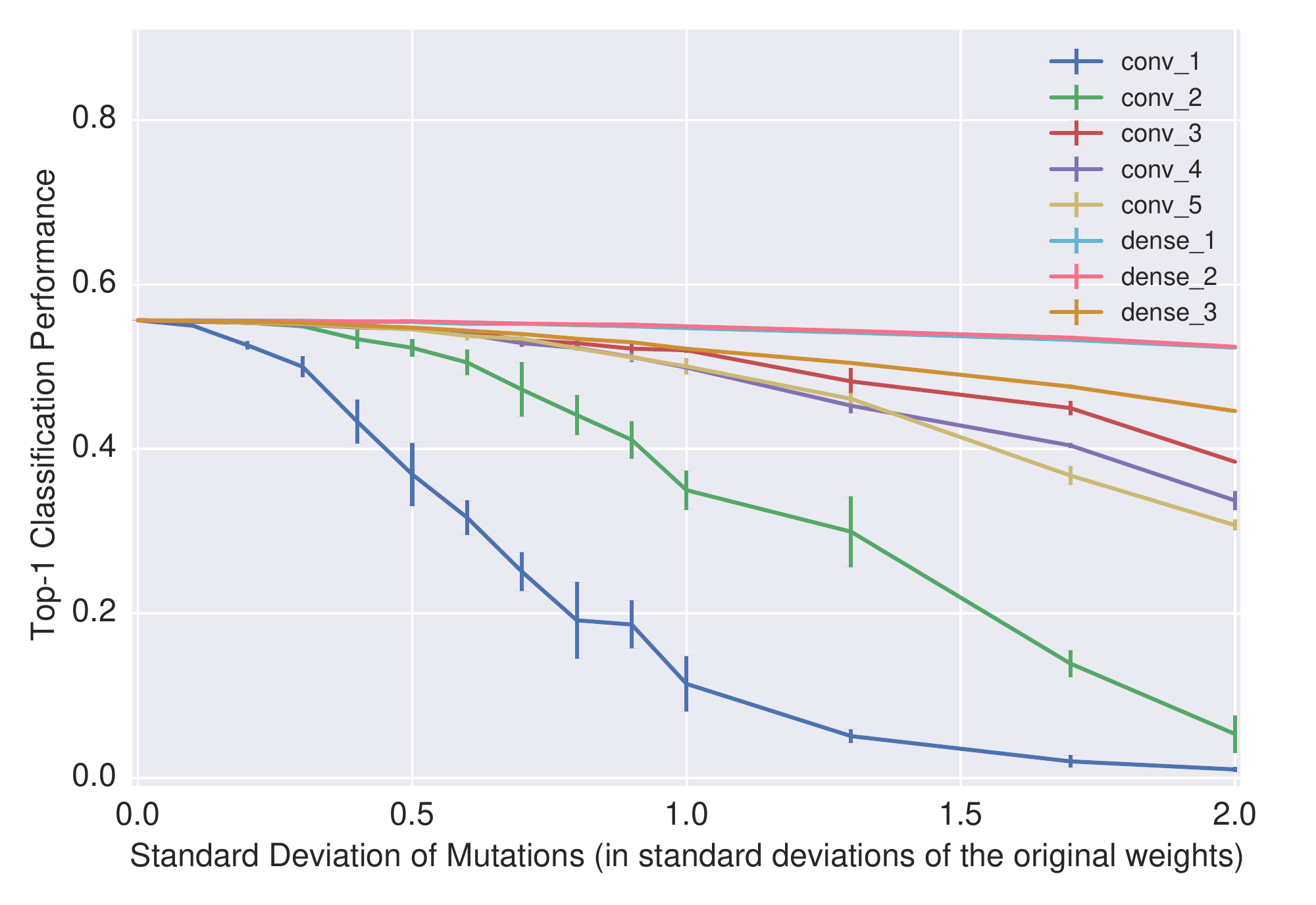}}
\caption{The top-1 classification performance of Alexnet on ILSVRC2012 after perturbing the weights with Gaussian mutations (Methods).  Note that the overall trends (performance of layers relative to one another, and drop-off in performance as weight mutation size increases) are consistent with the top-5 performance reported in Fig.~\ref{fig:perturbation:mutate-performance}, while the absolute performance metric on the $y$-axis differs.}
\label{fig:perturbation:mutate-performance-top1}
\end{center}
\vskip -0.2in
\end{figure}

Preliminary trials for other thresholds of classification strictness show qualitatively consistent results to the top-5 performance used throughout this paper -- with absolute performance metrics differing, but relational trends and qualitative results remaining consistent.  For example, the top-1 performance for Alexnet classification on ILSVRC2012 in Fig.~\ref{fig:perturbation:mutate-performance-top1} shows a very similar trend to Fig.~\ref{fig:perturbation:mutate-performance}.

\begin{figure}[ht]
\vskip 0.2in
\begin{center}
\centerline{
\includegraphics[width=\columnwidth]{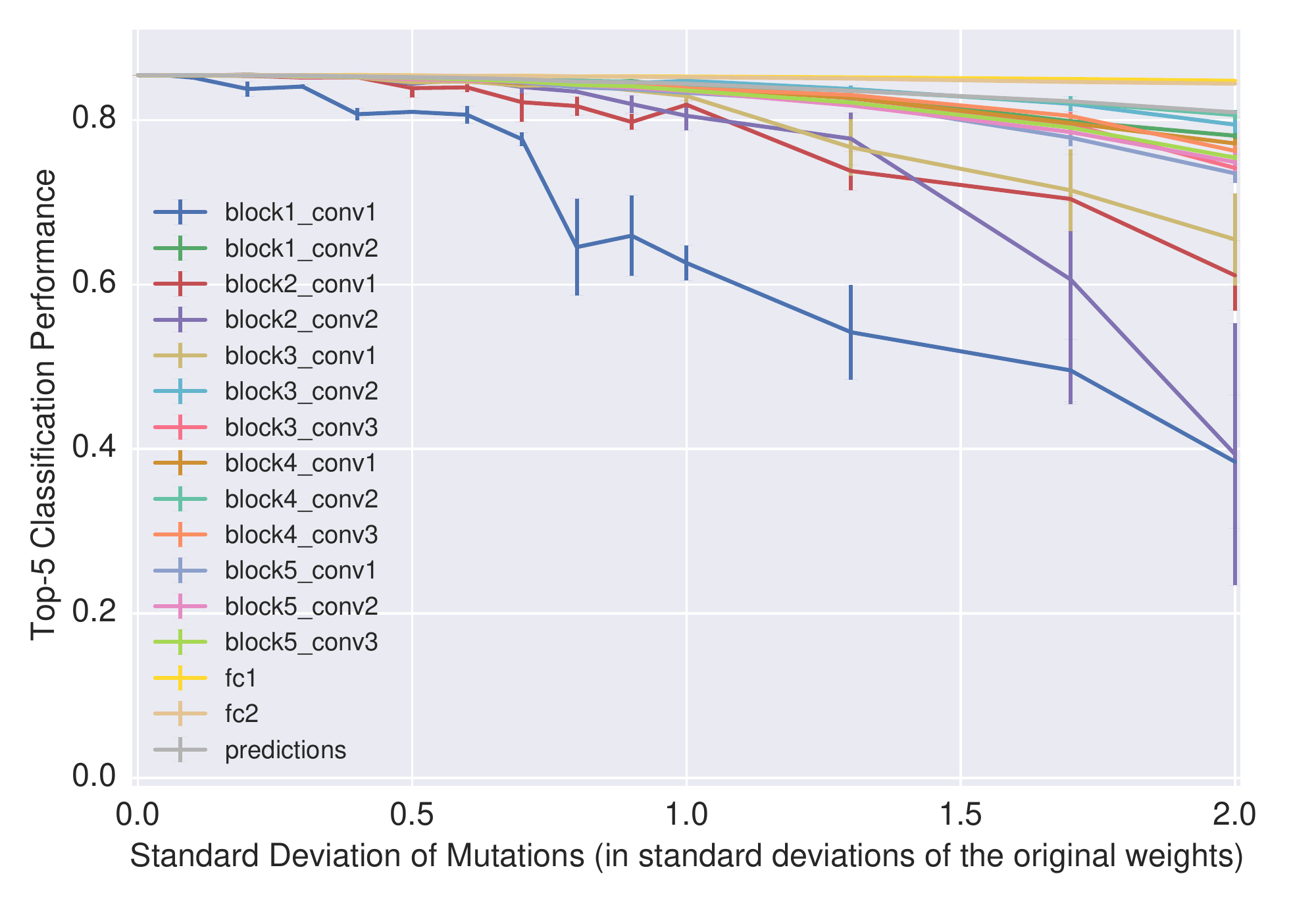}
}
\caption{VGG-16 top-5 performance on ILSVRC2012 after perturbing weights with Gaussian mutations.  Layers are listed in the order which they appear in the networks (those closer to the input layer first).  Note the similarities between the general trends on weight perturbations in this architecture, and those within the Alexnet architecture (Fig.~\ref{fig:perturbation:mutate-performance}).  Specifically, the drop off in performance as mutations size increases, yet the resilience to mutations in higher layers (e.g. all layers after block3\_conv1, show less than a 5.03\% drop in performance after a perturbation of magnitude 1, and a 23.9\% drop from their original performance ability after a mutation of magnitude 2).  We again observe latter layer in the networks showing more resilience to weight mutation than the earlier layers (with the notable exception of block1\_conv2, which shows surprising resilliance to mutation, losing an average of 3.56\% of its performance after mutations of magnitude 1 and 15.3\% from mutations of magnitude 2).  
}
\label{fig:perturbation:othernets}
\end{center}
\vskip -0.2in
\end{figure}

Preliminary results suggest that performance drop off is also qualitatively similar in other network architectures.  Fig.~\ref{fig:perturbation:othernets} shows a comparable effect of weight perturbations to the convolutional neural network VGG-16~\cite{simonyan2014very} which exhibits an increased network depth (16 layers compared to Alexnet's 8).

\section{Discussion}





The results from this work support our intuition of deep convolutional networks as robust function approximators -- and extends previous work on the robustness of networks classification towards perturbations of external stimuli to include an analysis of internal weight and architecture mutations.  

The convolutional networks examined here showed a significant degree of robustness to multiple forms of weight and topological alterations -- especially at the higher layers.   One could knock out over 70\% of synapses in any of the layers after conv\_2 of Alexnet and loose less than 30\% of its classification performance (Fig.~\ref{fig:perturbation:synapse-knockout}). Similarly, one could perturb every weight in those same layers by an average amount of weight value deviations in each respective layer, yet result in a drop in performance of less than 6\%.   

We also observed that perturbations tended to be more impactful when they targeted the first layer of the network.
Knocking out 30\% of the nodes in the first convolutional layer results in an average performance drop of over 70\%, and it only takes the removal of 50\% of nodes in this layer to loose over 95\% of the network's original classification performance.  


Changes to the first convolutional layer markedly differ from those to the upper layers, presenting a non-linear fall-off in classification performance.
Specifically, the removal of a single node from a largely-intact first convolutional layer results in a large fall-off to performance of the network -- while the removal of a node from an already sparse network has relatively little effect on performance.  This suggest that the filters in the first convolutional layer are codependent on one another, and the presence of one without another may provide relatively little information to the following layers.   

Turning to the case of synapse knockouts, we see that the drop-off in performance for the first convolutional layers is linearly dependent on the proportion of randomly removed synapses.  
However the relationship between the number of synapses knocked out and the effect on performance in the higher convolutional layers (2-5) appears nonlinear.  Specifically, the removal of a synapse has relatively little effect until a large proportion of synapses are removed -- after which each removed synapse then corresponds to a large performance drop.  This suggests that multiple instances of the information corresponding to each independent feature may be encoded in multiple different synapses.  Or it can be similarly interpreted to imply that nodes in these higher convolutional layers only rely on a subset of their incoming synapses to represent an independent feature.  


This seems to fit our intuition that the activation of high level features depend on the presence or absence of multiple (often redundant) lower level features.  Imagine the case where each node in a layer represents a single feature, and that feature is dependent on only one lower-level feature from the layer preceding it.  If this is the case, then we would see no difference in the classification performance after knocking out a given proportion of those lower-level features (nodes) and knocking out that same proportion of incoming connections to the higher layer features.  

However, the added robustness of networks to synapse knockouts (in comparison to node knockouts) suggests that any given feature in a deep convolutional networks (on average) relies on the presence of multiple lower-level features -- and that most of these lower-level features would be sufficient to provide correct classification in the absence of one of their \textquote{redundant features} (as evidenced by the finding that classification does not drop off linearly as the number of incoming connections to a higher-level features is randomly pruned).  There are, of course, a non-zero subset of images where one lower-level features provides information that is not contained as the classification performance monotomically drops for each progressive increase in synapse knockout proportion ($p \leq 0.0167$ for all increases in proportion, increasing from no knockouts to removing 10\% of synapses from layer dense\_2, for which $p=0.425$) -- meaning that the the information in any two features is unlikely to be perfectly redundant.  

The redundancy of deep convolutional networks has previously been evidenced by a robustness to different angles, sizes, or locations of objects in input images.  It could also be inferred by the architecture (size of subsequent layers) and connectivity (number of non-zero synapses leading into each node).  It also follows from intuition that higher-level layers compose features from lower-level layers in visual cortex~\cite{van1983hierarchical}.
However, the quantification of the degree to which each node -- on average -- is robust to removal of (or reliant upon the presence of) incoming information from lower-level features requires a knockout analysis of the internal architecture of the networks, such as the one provided in this work.


This explanation may also help to shed light onto the dichotomy between the first and higher-level convolutional layers, as the first layer (receiving only pixel information) may not have access to, or rely on, redundant incoming information in the way that higher convolutional layers do.  

It is also worth noting that part of the difference between perturbation impact on the lower and higher layers may also be due to the fact that information is fed forward sequentially through the network, such that  perturbations to each layer also affect the activation pattern fed into all the subsequent layers.  However, this effect would be expected to be dependent only on the number of layers above the perturbed one -- and our results show that the performance decreases tend to stagnate in the higher convolutional layers (3-5), suggesting that this effect alone does not account for the patterns observed above.   





The results stated above show the immediate impact of random perturbations on classification performance.  However, it is likely that convolutional networks undergoing online learning will also continue to adapt and recover from these undesirable changes.  Thus, while knowing the expected short term determent to performance is critical, we also plan to investigate the rate and maximum recoverability that would occur as a result of retraining in future work.

\section{Conclusion}

We examined the robustness of a pretrained convolutional neural network to internal weight and architecture perturbations.  We showed the classification performance effects of internal perturbations that removed nodes, removed synapses, and modified synapse weights.  We demonstrated that convolutional networks showed a significant degree of robustness to such changes. Perturbations to lower convolutional layers was significantly more impactful than perturbations to higher layers.  These results help us understand how information is encoded within the nodes and layers of deep convolutional neural networks.  

\bibliography{library}

\begin{thebibliography}{39}
\providecommand{\natexlab}[1]{#1}
\providecommand{\url}[1]{\texttt{#1}}
\expandafter\ifx\csname urlstyle\endcsname\relax
  \providecommand{\doi}[1]{doi: #1}\else
  \providecommand{\doi}{doi: \begingroup \urlstyle{rm}\Url}\fi

\bibitem[Bengio et~al.(2012)Bengio, Courville, and
  Vincent]{bengio2012unsupervised}
Bengio, Yoshua, Courville, Aaron~C, and Vincent, Pascal.
\newblock Unsupervised feature learning and deep learning: A review and new
  perspectives.
\newblock \emph{CoRR, abs/1206.5538}, 1, 2012.

\bibitem[Bottou \& LeCun(2003)Bottou and LeCun]{bottou2003large}
Bottou, L{\'e}on and LeCun, Yann.
\newblock Large scale online learning.
\newblock In \emph{NIPS}, volume~30, pp.\ ~77, 2003.

\bibitem[Chiel \& Beer(1997)Chiel and Beer]{chiel1997brain}
Chiel, Hillel~J and Beer, Randall~D.
\newblock The brain has a body: adaptive behavior emerges from interactions of
  nervous system, body and environment.
\newblock \emph{Trends in neurosciences}, 20\penalty0 (12):\penalty0 553--557,
  1997.

\bibitem[Chollet(2015)]{Chollet2015}
Chollet, Fran{\c{c}}ois.
\newblock {Keras}, 2015.
\newblock URL \url{https://github.com/fchollet/keras}.

\bibitem[Cunningham(1982)]{cunningham1982naturally}
Cunningham, Timothy~J.
\newblock Naturally occurring neuron death and its regulation by developing
  neural pathways.
\newblock \emph{International review of cytology}, 74:\penalty0 163--186, 1982.

\bibitem[Deng et~al.(2009)Deng, Dong, Socher, Li, Li, and
  Fei-Fei]{deng2009imagenet}
Deng, Jia, Dong, Wei, Socher, Richard, Li, Li-Jia, Li, Kai, and Fei-Fei, Li.
\newblock Imagenet: A large-scale hierarchical image database.
\newblock In \emph{Computer Vision and Pattern Recognition, 2009. CVPR 2009.
  IEEE Conference on}, pp.\  248--255. IEEE, 2009.

\bibitem[Eriksson et~al.(1998)Eriksson, Perfilieva, Bj{\"o}rk-Eriksson, Alborn,
  Nordborg, Peterson, and Gage]{eriksson1998neurogenesis}
Eriksson, Peter~S, Perfilieva, Ekaterina, Bj{\"o}rk-Eriksson, Thomas, Alborn,
  Ann-Marie, Nordborg, Claes, Peterson, Daniel~A, and Gage, Fred~H.
\newblock Neurogenesis in the adult human hippocampus.
\newblock \emph{Nature medicine}, 4\penalty0 (11):\penalty0 1313--1317, 1998.

\bibitem[Goodfellow et~al.(2014)Goodfellow, Shlens, and
  Szegedy]{Goodfellow2014}
Goodfellow, Ian~J, Shlens, Jonathon, and Szegedy, Christian.
\newblock Explaining and harnessing adversarial examples.
\newblock \emph{arXiv preprint arXiv:1412.6572}, 2014.

\bibitem[Hebb(2005)]{hebb2005organization}
Hebb, Donald~Olding.
\newblock \emph{The organization of behavior: A neuropsychological theory}.
\newblock Psychology Press, 2005.

\bibitem[Hinton et~al.(2012)Hinton, Srivastava, Krizhevsky, Sutskever, and
  Salakhutdinov]{hinton2012improving}
Hinton, Geoffrey~E, Srivastava, Nitish, Krizhevsky, Alex, Sutskever, Ilya, and
  Salakhutdinov, Ruslan~R.
\newblock Improving neural networks by preventing co-adaptation of feature
  detectors.
\newblock \emph{arXiv preprint arXiv:1207.0580}, 2012.

\bibitem[Hornik et~al.(1989)Hornik, Stinchcombe, and
  White]{hornik1989multilayer}
Hornik, Kurt, Stinchcombe, Maxwell, and White, Halbert.
\newblock Multilayer feedforward networks are universal approximators.
\newblock \emph{Neural networks}, 2\penalty0 (5):\penalty0 359--366, 1989.

\bibitem[Huval et~al.(2015)Huval, Wang, Tandon, Kiske, Song, Pazhayampallil,
  Andriluka, Rajpurkar, Migimatsu, Cheng-Yue, et~al.]{huval2015empirical}
Huval, Brody, Wang, Tao, Tandon, Sameep, Kiske, Jeff, Song, Will,
  Pazhayampallil, Joel, Andriluka, Mykhaylo, Rajpurkar, Pranav, Migimatsu,
  Toki, Cheng-Yue, Royce, et~al.
\newblock An empirical evaluation of deep learning on highway driving.
\newblock \emph{arXiv preprint arXiv:1504.01716}, 2015.

\bibitem[Jabri \& Flower(1992)Jabri and Flower]{jabri1992weight}
Jabri, Marwan and Flower, Barry.
\newblock Weight perturbation: An optimal architecture and learning technique
  for analog vlsi feedforward and recurrent multilayer networks.
\newblock \emph{IEEE Transactions on Neural Networks}, 3\penalty0 (1):\penalty0
  154--157, 1992.

\bibitem[Jia et~al.(2014)Jia, Shelhamer, Donahue, Karayev, Long, Girshick,
  Guadarrama, and Darrell]{jia2014caffe}
Jia, Yangqing, Shelhamer, Evan, Donahue, Jeff, Karayev, Sergey, Long, Jonathan,
  Girshick, Ross, Guadarrama, Sergio, and Darrell, Trevor.
\newblock Caffe: Convolutional architecture for fast feature embedding.
\newblock In \emph{Proceedings of the 22nd ACM international conference on
  Multimedia}, pp.\  675--678. ACM, 2014.

\bibitem[Krizhevsky et~al.(2012)Krizhevsky, Sutskever, and
  Hinton]{krizhevsky2012imagenet}
Krizhevsky, Alex, Sutskever, Ilya, and Hinton, Geoffrey~E.
\newblock Imagenet classification with deep convolutional neural networks.
\newblock In \emph{Advances in neural information processing systems}, pp.\
  1097--1105, 2012.

\bibitem[LeCun et~al.(2015)LeCun, Bengio, and Hinton]{lecun2015deep}
LeCun, Yann, Bengio, Yoshua, and Hinton, Geoffrey.
\newblock Deep learning.
\newblock \emph{Nature}, 521\penalty0 (7553):\penalty0 436--444, 2015.

\bibitem[LeCun et~al.(2012)LeCun, Bottou, Orr, and
  M{\"u}ller]{lecun2012efficient}
LeCun, Yann~A, Bottou, L{\'e}on, Orr, Genevieve~B, and M{\"u}ller,
  Klaus-Robert.
\newblock Efficient backprop.
\newblock In \emph{Neural networks: Tricks of the trade}, pp.\  9--48.
  Springer, 2012.

\bibitem[Lee et~al.(2009)Lee, Grosse, Ranganath, and Ng]{lee2009convolutional}
Lee, Honglak, Grosse, Roger, Ranganath, Rajesh, and Ng, Andrew~Y.
\newblock Convolutional deep belief networks for scalable unsupervised learning
  of hierarchical representations.
\newblock In \emph{Proceedings of the 26th annual international conference on
  machine learning}, pp.\  609--616. ACM, 2009.

\bibitem[Levine et~al.(2016)Levine, Finn, Darrell, and Abbeel]{levine2016end}
Levine, Sergey, Finn, Chelsea, Darrell, Trevor, and Abbeel, Pieter.
\newblock End-to-end training of deep visuomotor policies.
\newblock \emph{Journal of Machine Learning Research}, 17\penalty0
  (39):\penalty0 1--40, 2016.

\bibitem[Liao \& Yu(1998)Liao and Yu]{liao1998robust}
Liao, Xiaofeng and Yu, Jeubang.
\newblock Robust stability for interval hopfield neural networks with time
  delay.
\newblock \emph{IEEE Transactions on Neural Networks}, 9\penalty0 (5):\penalty0
  1042--1045, 1998.

\bibitem[Mead \& Ismail(2012)Mead and Ismail]{mead2012analog}
Mead, Carver and Ismail, Mohammed.
\newblock \emph{Analog VLSI implementation of neural systems}, volume~80.
\newblock Springer Science \& Business Media, 2012.

\bibitem[Moody et~al.(1991)]{Moody1991}
Moody, John~E et~al.
\newblock The effective number of parameters: An analysis of generalization and
  regularization in nonlinear learning systems.
\newblock In \emph{NIPS}, volume~4, pp.\  847--854, 1991.

\bibitem[Nguyen et~al.(2015)Nguyen, Yosinski, and Clune]{Nguyen2015}
Nguyen, Anh, Yosinski, Jason, and Clune, Jeff.
\newblock Deep neural networks are easily fooled: High confidence predictions
  for unrecognizable images.
\newblock In \emph{Proceedings of the IEEE Conference on Computer Vision and
  Pattern Recognition}, pp.\  427--436, 2015.

\bibitem[Ross et~al.(2013)Ross, Melik-Barkhudarov, Shankar, Wendel, Dey,
  Bagnell, and Hebert]{ross2013learning}
Ross, St{\'e}phane, Melik-Barkhudarov, Narek, Shankar, Kumar~Shaurya, Wendel,
  Andreas, Dey, Debadeepta, Bagnell, J~Andrew, and Hebert, Martial.
\newblock Learning monocular reactive uav control in cluttered natural
  environments.
\newblock In \emph{Robotics and Automation (ICRA), 2013 IEEE International
  Conference on}, pp.\  1765--1772. IEEE, 2013.

\bibitem[Russakovsky et~al.(2015)Russakovsky, Deng, Su, Krause, Satheesh, Ma,
  Huang, Karpathy, Khosla, Bernstein, et~al.]{russakovsky2015imagenet}
Russakovsky, Olga, Deng, Jia, Su, Hao, Krause, Jonathan, Satheesh, Sanjeev, Ma,
  Sean, Huang, Zhiheng, Karpathy, Andrej, Khosla, Aditya, Bernstein, Michael,
  et~al.
\newblock Imagenet large scale visual recognition challenge.
\newblock \emph{International Journal of Computer Vision}, 115\penalty0
  (3):\penalty0 211--252, 2015.

\bibitem[Salakhutdinov \& Hinton(2009)Salakhutdinov and
  Hinton]{salakhutdinov2009deep}
Salakhutdinov, Ruslan and Hinton, Geoffrey~E.
\newblock Deep boltzmann machines.
\newblock In \emph{AISTATS}, volume~1, pp.\ ~3, 2009.

\bibitem[Schneider(2017)]{schneider2017deeper}
Schneider, David.
\newblock Deeper and cheaper machine learning [top tech 2017].
\newblock \emph{IEEE Spectrum}, 54\penalty0 (1):\penalty0 42--43, 2017.

\bibitem[Simonyan \& Zisserman(2014)Simonyan and Zisserman]{simonyan2014very}
Simonyan, Karen and Zisserman, Andrew.
\newblock Very deep convolutional networks for large-scale image recognition.
\newblock \emph{arXiv preprint arXiv:1409.1556}, 2014.

\bibitem[Song et~al.(2000)Song, Miller, and Abbott]{song2000competitive}
Song, Sen, Miller, Kenneth~D, and Abbott, Larry~F.
\newblock Competitive hebbian learning through spike-timing-dependent synaptic
  plasticity.
\newblock \emph{Nature neuroscience}, 3\penalty0 (9):\penalty0 919--926, 2000.

\bibitem[Srivastava et~al.(2014{\natexlab{a}})Srivastava, Hinton, Krizhevsky,
  Sutskever, and Salakhutdinov]{Srivastava2014}
Srivastava, Nitish, Hinton, Geoffrey, Krizhevsky, Alex, Sutskever, Ilya, and
  Salakhutdinov, Ruslan.
\newblock {Dropout: A Simple Way to Prevent Neural Networks from Overfitting}.
\newblock \emph{Journal of Machine Learning Research}, 15:\penalty0 1929--1958,
  2014{\natexlab{a}}.

\bibitem[Srivastava et~al.(2014{\natexlab{b}})Srivastava, Hinton, Krizhevsky,
  Sutskever, and Salakhutdinov]{srivastava2014dropout}
Srivastava, Nitish, Hinton, Geoffrey~E, Krizhevsky, Alex, Sutskever, Ilya, and
  Salakhutdinov, Ruslan.
\newblock Dropout: a simple way to prevent neural networks from overfitting.
\newblock \emph{Journal of Machine Learning Research}, 15\penalty0
  (1):\penalty0 1929--1958, 2014{\natexlab{b}}.

\bibitem[Team et~al.(2016)Team, Al-Rfou, Alain, Almahairi, Angermueller,
  Bahdanau, Ballas, Bastien, Bayer, Belikov, et~al.]{team2016theano}
Team, The Theano~Development, Al-Rfou, Rami, Alain, Guillaume, Almahairi,
  Amjad, Angermueller, Christof, Bahdanau, Dzmitry, Ballas, Nicolas, Bastien,
  Fr{\'e}d{\'e}ric, Bayer, Justin, Belikov, Anatoly, et~al.
\newblock Theano: A python framework for fast computation of mathematical
  expressions.
\newblock \emph{arXiv preprint arXiv:1605.02688}, 2016.

\bibitem[Van~Essen \& Maunsell(1983)Van~Essen and
  Maunsell]{van1983hierarchical}
Van~Essen, David~C and Maunsell, John~HR.
\newblock Hierarchical organization and functional streams in the visual
  cortex.
\newblock \emph{Trends in neurosciences}, 6:\penalty0 370--375, 1983.

\bibitem[Widrow \& Lehr(1990)Widrow and Lehr]{widrow199030}
Widrow, Bernard and Lehr, Michael~A.
\newblock 30 years of adaptive neural networks: perceptron, madaline, and
  backpropagation.
\newblock \emph{Proceedings of the IEEE}, 78\penalty0 (9):\penalty0 1415--1442,
  1990.

\bibitem[Wilcoxon(1945)]{wilcoxon1945individual}
Wilcoxon, Frank.
\newblock Individual comparisons by ranking methods.
\newblock \emph{Biometrics bulletin}, 1\penalty0 (6):\penalty0 80--83, 1945.

\bibitem[Yim et~al.(2007)Yim, Shen, Salemi, Rus, Moll, Lipson, Klavins, and
  Chirikjian]{yim2007modular}
Yim, Mark, Shen, Wei-Min, Salemi, Behnam, Rus, Daniela, Moll, Mark, Lipson,
  Hod, Klavins, Eric, and Chirikjian, Gregory~S.
\newblock Modular self-reconfigurable robot systems [grand challenges of
  robotics].
\newblock \emph{IEEE Robotics \& Automation Magazine}, 14\penalty0
  (1):\penalty0 43--52, 2007.

\bibitem[Zeiler \& Fergus(2014)Zeiler and Fergus]{zeiler2014visualizing}
Zeiler, Matthew~D and Fergus, Rob.
\newblock Visualizing and understanding convolutional networks.
\newblock In \emph{European conference on computer vision}, pp.\  818--833.
  Springer, 2014.

\bibitem[Zhang et~al.(2016)Zhang, Bengio, Hardt, Recht, and Vinyals]{Zhang2016}
Zhang, Chiyuan, Bengio, Samy, Hardt, Moritz, Recht, Benjamin, and Vinyals,
  Oriol.
\newblock Understanding deep learning requires rethinking generalization.
\newblock \emph{arXiv preprint arXiv:1611.03530}, 2016.

\bibitem[Zurada et~al.(1997)Zurada, Malinowski, and
  Usui]{zurada1997perturbation}
Zurada, Jacek~M, Malinowski, Aleksander, and Usui, Shiro.
\newblock Perturbation method for deleting redundant inputs of perceptron
  networks.
\newblock \emph{Neurocomputing}, 14\penalty0 (2):\penalty0 177--193, 1997.

\end{thebibliography}
\bibliographystyle{icml2017}

\begin{table*}
\begin{center}
\def\arraystretch{1.2}
\begin{tabular}{ | l | c | c | c | c | c | c | c | c | }
  \hline
  layer & size & mean & median & $\sigma$ & min & max &kurtosis & skew \\ \hline
conv\_1\_W & 34848 & 2.84e-06 & 0.000245 & 0.0559 & -0.421 & 0.416 & 4.93 & -0.0331 \\ \hline
conv\_1\_b & 96 & -0.476 & -0.437 & 0.452 & -1.59 & 0.321 & -0.39 & -0.422 \\ \hline
conv\_2\_1\_W & 153600 & -0.00107 & -0.00174 & 0.0261 & -0.21 & 0.637 & 11.9 & 1.06 \\ \hline
conv\_2\_1\_b & 128 & 0.104 & 0.104 & 0.025 & -0.0153 & 0.179 & 4.96 & -1.02 \\ \hline
conv\_2\_2\_W & 153600 & -0.000902 & -0.00108 & 0.0294 & -0.266 & 0.45 & 8.79 & 0.631 \\ \hline
conv\_2\_2\_b & 128 & 0.109 & 0.107 & 0.0294 & 0.0282 & 0.207 & 0.95 & 0.51 \\ \hline
conv\_3\_W & 884736 & -3.91e-05 & -0.000774 & 0.0182 & -0.23 & 0.606 & 8.98 & 0.818 \\ \hline
conv\_3\_b & 384 & 0.0105 & 0.00943 & 0.0384 & -0.136 & 0.183 & 1.47 & 0.137 \\ \hline
conv\_4\_1\_W & 331776 & -0.000262 & -0.00159 & 0.0192 & -0.191 & 0.457 & 4.54 & 0.73 \\ \hline
conv\_4\_1\_b & 192 & 0.123 & 0.125 & 0.0561 & -0.0196 & 0.276 & 0.121 & -0.122 \\ \hline
conv\_4\_2\_W & 331776 & -0.000504 & -0.0019 & 0.0222 & -0.193 & 0.45 & 3.4 & 0.653 \\ \hline
conv\_4\_2\_b & 192 & 0.124 & 0.13 & 0.0724 & -0.0983 & 0.29 & 0.358 & -0.517 \\ \hline
conv\_5\_1\_W & 221184 & -0.0021 & -0.00408 & 0.0226 & -0.17 & 0.382 & 5.68 & 1.08 \\ \hline
conv\_5\_1\_b & 128 & 0.0462 & 0.0295 & 0.0951 & -0.167 & 0.347 & 1.11 & 0.941 \\ \hline
conv\_5\_2\_W & 221184 & -0.00231 & -0.00471 & 0.026 & -0.2 & 0.336 & 5.04 & 1.08 \\ \hline
conv\_5\_2\_b & 128 & 0.0282 & 0.021 & 0.0894 & -0.202 & 0.262 & 0.0409 & 0.377 \\ \hline
dense\_1\_W & 37748736 & -9.88e-05 & -0.000156 & 0.00229 & -0.0174 & 0.0264 & 0.587 & 0.174 \\ \hline
dense\_1\_b & 4096 & 0.0516 & 0.0516 & 0.00522 & 0.0336 & 0.0731 & 0.236 & -0.00771 \\ \hline
dense\_2\_W & 16777216 & -0.000239 & -0.000396 & 0.0032 & -0.0175 & 0.0377 & 0.187 & 0.284 \\ \hline
dense\_2\_b & 4096 & 0.09 & 0.0887 & 0.02 & 0.0343 & 0.183 & 0.226 & 0.388 \\ \hline
dense\_3\_W & 4096000 & -2.42e-07 & -0.000845 & 0.00581 & -0.0267 & 0.0422 & 0.271 & 0.558 \\ \hline
dense\_3\_b & 1000 & -5.96e-06 & 0.00228 & 0.0749 & -0.208 & 0.235 & -0.105 & 0.0442 \\ \hline
\end{tabular}
\caption{ Descriptive statistics for the unperturbed Alexnet pretrained on ILSVRC2012.  Layer naming conventions consist of the following properties of the layer (in order and separated by an underscore): the type of layer (convolutional or fully-connected dense layer), the order that the layer appears within its layer type (rows are also ordered within the layer the appear in the network, with rows appearing first in the table being closer to the input layer of the network), the second number represents the split layer order (if only one numeric value appears, then the layer is not split), and finally the type of parameter encoded in the layer (synapse weight or node bias). 
Note that the synapse weight and node bias parameters are implemented separately in Alexnet, and for the calculating the magnitude of perturbations in the methods above, we use different standard deviation values ($\sigma$) for the weight and bias in each layers, as these values may differ.  Weight mutations were chosen to be normalized by standard deviation of the original parameter values in order to keep the relative \textquote{impact} of mutations consistent across layers, as the standard deviation between parameters in different layers may differ my up to two orders of magnitude.  When a node is removed in the node-knockout treatment, its bias parameter is also set to zero.  Similarly, bias parameters are randomly selected to be set to zero in the same way as synapses for the synapse-knockout treatment.}
\label{alexnetTable}
\end{center}
\end{table*}

\end{document}


\twocolumn[
\icmltitle{Robustness of neural networks \\ 
           to weight changes}

\icmlauthor{Your Name}{email@yourdomain.edu}
\icmladdress{Your Fantastic Institute,
            314159 Pi St., Palo Alto, CA 94306 USA}
\icmlauthor{Your CoAuthor's Name}{email@coauthordomain.edu}
\icmladdress{Their Fantastic Institute,
            27182 Exp St., Toronto, ON M6H 2T1 CANADA}

\icmlkeywords{boring formatting information, machine learning, ICML}

\vskip 0.3in
]

\begin{figure}[ht]
\vskip 0.2in
\begin{center}
\centerline{\includegraphics[width=\columnwidth]{figures/perturbations_draw-performance_ILSVRC2012val.pdf}}
\caption{Performance on the ILSVRC2012 validation set after perturbing the weights with random draws from the distribution of a layer (Methods)}
\label{fig:perturbation:draw-performance}
\end{center}
\vskip -0.2in
\end{figure}

\lipsum[66]

\bibliography{library}
\bibliographystyle{icml2016}